# A FAST AND PRECISE METHOD FOR LARGE-SCALE LAND-USE MAPPING BASED ON DEEP LEARNING

*Xuan Yang[1, 3], Zhengchao Chen[2], Baipeng Li[2], Dailiang Peng[1], Pan Chen[2, 3], Bing Zhang[1, 3, ]\**

[1]Key Laboratory of Digital Earth Science, Institute of Remote Sensing and Digital Earth, Chinese Academy of Sciences, Beijing 100094, China
[2]Airborne Remote Sensing Center, Institute of Remote Sensing and Digital Earth, Chinese Academy of Sciences, Beijing 100094, China
[3]University of Chinese Academy of Sciences, Beijing 100049, China

## ABSTRACT

The land-use map is an important data that can reflect the use and transformation of human land, and can provide valuable reference for land-use planning. For the traditional image classification method, producing a high spatial resolution (HSR), land-use map in large-scale is a big project that requires a lot of human labor, time, and financial expenditure. The rise of the deep learning technique provides a new solution to the problems above. This paper proposes a fast and precise method that can achieve large-scale land-use classification based on deep convolutional neural network (DCNN). In this paper, we optimize the data tiling method and the structure of DCNN for the multi-channel data and the splicing edge effect, which are unique to remote sensing deep learning, and improve the accuracy of land-use classification. We apply our improved methods in the Guangdong Province of China using GF-1 images, and achieve the land-use classification accuracy of 81.52%. It takes only 13 hours to complete the work, which will take several months for human labor.

*Index Terms—* Land-use Mapping, Deep Learning, Big Data, Semantic Segmentation

## 1. INTRODUCTION

The land-use map is an important data product. It shows the utilization of land resources and the transformation results of human beings. It reflects the land-use form and functional use, and has important reference value for the overall planning of land-use.

The traditional image classification is based on the artificially designed feature extractor. There are many contributions based on per-pixel, object-based and per-field classification method [1-6]. In the era of remote sensing big data, the traditional classification model has weak generalization ability and no universality. Even the land-use map based on traditional methods still needs to be manually corrected.

The deep learning technique is a data-driven approach that can extract rich information from big data. Deep learning uses DCNN structure, which contains a large number of learnable parameters, and automatically learns network parameters through supervised learning [7]. In the era of remote sensing big data, using the deep learning technique to replace the traditional method can greatly reduce the consumption of human labor, time and financial expenditure, and can obtain a large-scale land-use map product with higher precision.

In recent years, the deep learning technique has made rapid progress in the field of computer vision. In 2012, Alex Krizhevsky et al. won the ImageNet Large Scale Visual Recognition Challenge (ILSVRC) by using AlexNet [8], and DCNN began to be widely used in image classification. There are more DCNN architecture based on fully convolutional neural network has been designed for semantic segmentation. UNet and SegNet are designed with Encode-Decode structure [9, 10], PSPNet and DeepLab are designed with dilated convolution [11, 12].

Thanks to the development of deep learning in the field of computer vision, DCNN can be transferred into the field of remote sensing. Most of DCNN are designed for simple photographs, which are composed of only 3 channels of RGB data. Therefore, the transferred DCNN cannot process remote sensing images, which have more than 3 channels of data. In the field of computer vision, the simple photographs are independent with each other, the size is relative small, and the scene is relative simple. However, remote sensing images are geographically related, with large size and complex scenes. Due to hardware limitations, remote sensing images cannot be fed into DCNN directly, and additional slicing is required, which leads to the unique edge effect of remote sensing deep learning.

In this study, the Guangdong Province is selected as the study area, and Chinese GF-1 satellite images are used for land-use mapping. We adopt the semantic segmentation network PSPNet as feature extractor and classifier for land-use map. All of four channels of GF-1 data, Normalized Difference Vegetation Index (NDVI) and Normalized Difference Water Index (NDWI) data will be fed into PSPNet, and the image slicing method with overlapping regions will be used to reduce and eliminate the edge effect.

## 2. METHODOLOGY

### 2.1. Data and Classification System

Chinese GF-1 satellite images at the spatial resolution of 8 m in 2017, over the Guangdong Province with the area of 179,700 km$^2$, are obtained. We perform the necessary preprocessing on the GF-1 image through Pixel Factory. Finally, 466 original images are available for training and inference. Data volume is up to 269 GB. The preprocessed GF-1 image consists of 4 channels of data (B, G, R, NIR) with spatial resolution of 2 m, single image size of 15200 × 10200 pixels, and adjacent images with 200 pixels overlapped. The GF-1 images of the Guangdong Province are shown in Fig.1 (a).

The land-use classification label data was labeled in 2015 by artificial interpretation and field survey. There are 188 classification maps that can be used as the training and validation sets, which cover 40.34% of the Guangdong Province with an area of 72,500 km$^2$. Single label size is 15001 × 10001 pixels and the spatial resolution is 2 m. Data volume is up to 118 GB. The land-use classification label data is shown in Fig.1 (b).

The classification system used in this paper is determined by the sample label data and the spatial resolution of the original image. There are 9 categories in classification label data: cultivated land, garden land, forest land, grass land, water body, residential area, road, bare land, agricultural facilities. All of the categories can be identified by DCNN in 2 m spatial resolution, and are finally adopted as the classification system of land-use map in the Guangdong Province of China.

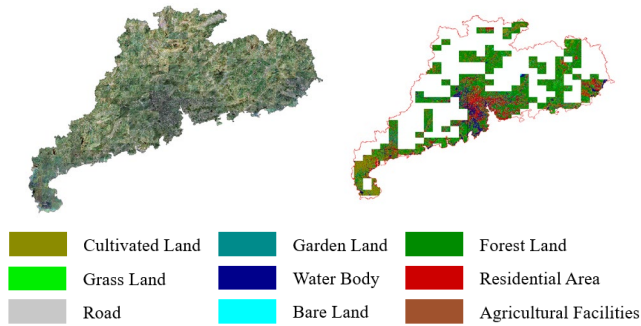

(a) GF-1 images of Guangdong (left)
(b) Land-use classification label (right)
Fig.1 Source Data

### 2.2. Data Slicing

The sample data size far exceeds the load capacity of the GPU memory, so the sample data needs to be sliced. A larger scale feature can be learned by DCNN with a larger slice image. This paper slices the sample data into 640 × 640 pixels. The remote sensing image slice makes the features on the image boundary incomplete thereby reducing the accuracy of the tile image boundary. The same kind of objects that cross two adjacent slices may be identified as two different objects. Therefore, in this paper, a 50% overlap region is set for image slicing, in another words, there are 320 pixels overlapping between adjacent tiles.

In order to maximize the sample utilization rate, the sample data is padded by 160 pixels so that the boundary region of sample data can also be in the central region of the DCNN feature map. In this paper, we flip the normal image block to the padded image block, making the padding area closer to the real situation and eliminating the side effects of sharp gradient change between the normal image block and the padded image block, which is a problem if padding zero value to the padded image block.

The training set includes 179 GF-1 images. Using the slicing method that mentioned above, 206,492 tiles are obtained as the final training data. 5% of the sample data (9 GF-1 images) was selected to evaluate the accuracy of the DCNN model and not be fed into network in the training.

### 2.3. Training and Inference

We modify the PSPNet structure as the DCNN model for training and inference so that it can be fed 6-channel data, including blue, green, red, infrared, NDVI, and NDWI data. In order to avoid over-fitting of parameters, random data augment is performed on the input image.

In the training stage, we use the parameters of ImageNet's pre-training model to initialize and fine-tune ResNet-50. It can speed up network learning and improve model accuracy. In this paper, PSPNet is implemented based on PyTorch deep learning framework. The optimizer is Adam with the learning rate of 1e-5, the weight decay of 5e-4, the momentum of 0.99 and the batch size of 12, running on NVIDIA TITAN XP with 12 GB GPU memory. It takes 4 days to train the training set of 206,492 tiles with data volume of 394 GB.

In the inference stage, it is also necessary to slice the original data using the slicing method mentioned above. The PSPNet based on PyTorch deep learning framework is the same as the training phase with the batch size of 60. It takes 13 hours to infer the land-use map of the whole Guangdong Province. The inference data volume reaches 67.3 GB, and the coverage area is 179,700 km$^2$.

## 3. RESULTS AND ANALYSIS

Land-use map in the Guangdong Province of China based on deep learning is shown in Fig.2. From the classification result, it can be seen intuitively that the forest land accounts for most of the area of the Guangdong Province. The residential areas are mainly distributed on the Pearl River Delta, the coastal areas of eastern Guangdong and western Guangdong. The Leizhou Peninsula is dominated by cultivated land. The overall classification results are basically consistent with the facts. The samples used for DCNN training accounted for 40.34% of all samples. The model does not over-fit and the generalization ability is strong.

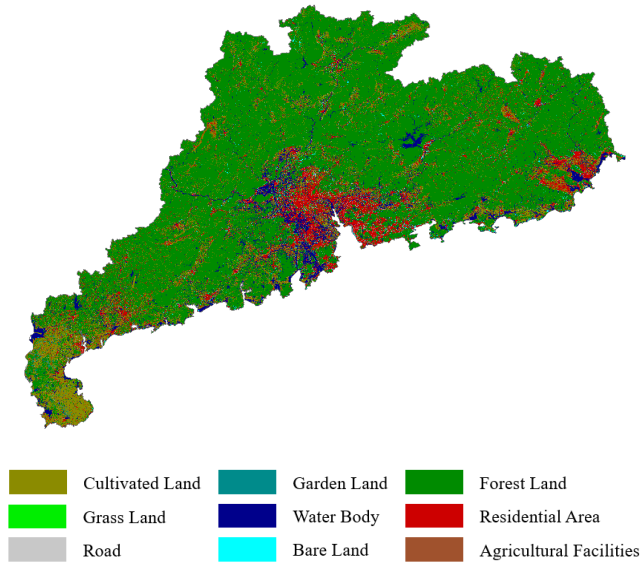

Fig.2 Land-use Map of the Guangdong Province

In this paper, two models are trained, one for 3-channel images (B, G, R) and the other for 6-channel images (B, G, R, NIR, NDVI, NDWI) for comparison. Fig.3 shows the specific details of the classification results. Among them, Column a (original image) are pseudo color composite images that are fused with NIR, R, and G channel of GF-1 satellite imagery. Column b (ground truth) are the classification label data. Column c (Land-use 3, short for LU-3 hereafter) are the results that are inferred by 3-channel input training model. Column d (Land-use 6, short for LU-6 hereafter) are inferred by 6-channel input training model.

Fig.3 (1a) is a village with a road passing through the middle of the village. As shown in Fig.3 (1c), this road is basically unrecognized and covered by surrounding residential areas. However, the road is correctly classified in LU-6. Multi-channel data can enhance the feature difference of road and residential area to better distinguish between these two categories. The bare land in the ground truth in Fig.3 (2a) does not match the original image. It is possible that there was bare land during the field survey, but was changed to other uses later. This suggests that the original image and the label data are not from the same year, therefore the problem is inevitable. There is a tall building in the center of the original image, next to the shadow of the building. The shadow is classified as a water body in LU-3. This shows that when there is no NIR channel data, the shadow is easily confused with the water body.

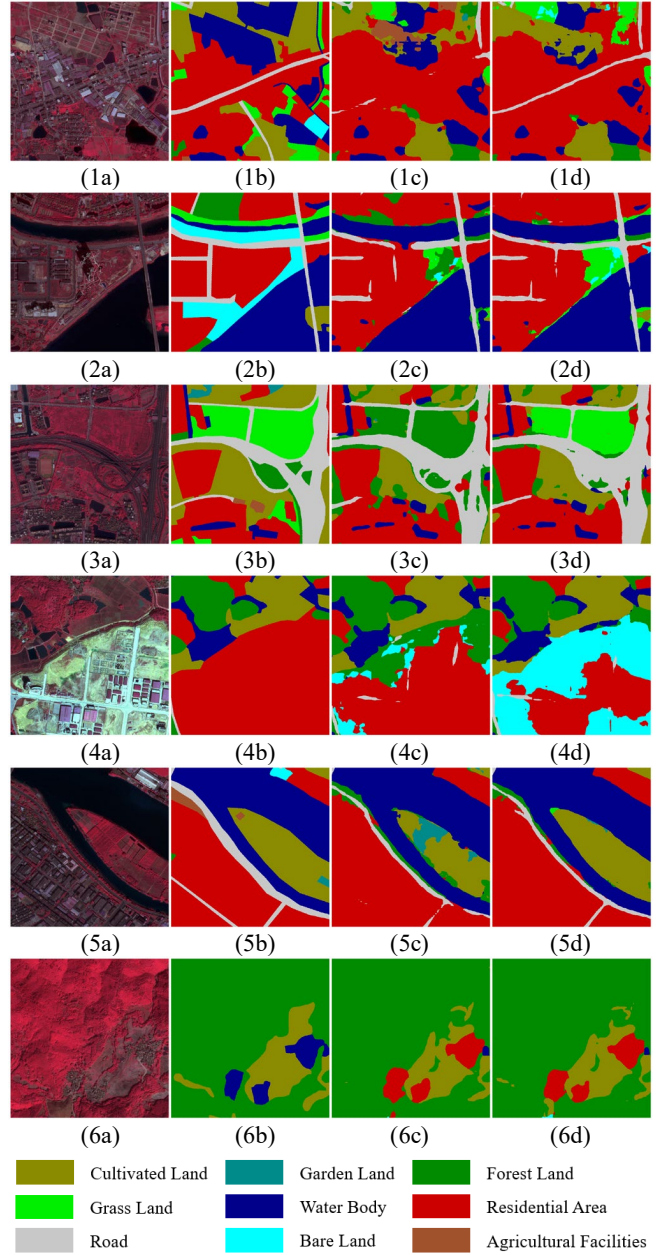

Fig.3 Details of the classification results, where column (a) are original images; column (b) are classification label data; column (c) are results of 3-channel input model; column (d) are results of 6-channel input model.

Fig.3 (3a) shows that there is the overpass of the expressway with a grass land next to it. This grass land is classified as forest land in LU-3. This is because the NDVI of grass land and forest land is different from each other. With the help of the NDVI channel, LU-6 can distinguish them. The lower right corner of Fig.3 (4a) is an industrial area surrounded by bare land. The whole area is labeled as a residential area in Fig.3 (4b). In LU-3, this area is basically classified as a residential area, but part of this area is classified as grass land and bare land. The error is obvious in LU-3. However, this area is classified correctly in LU-6. This indicates that bare land is easily confused with residential areas. In Fig.3 (5a), there is a small island in the middle of the river. The island is mainly cultivated land. It is classified correctly in LU-6, but confused with garden land and grass land in LU-3. This is because without the NDVI channel data, cultivated land looks like garden land and grass land in RGB true color image. Fig.3 (6a) shows three small villages. They are incorrectly labeled as water bodies in ground truth as shown in Fig.3 (6b), however, they are classified as residential areas in LU-3 and LU-6, matching the original image. It shows that the deep learning technique can tolerate a small number of incorrect labels in the sample through self-learning, which shows that the deep learning technique has certain reliability.

The accuracy evaluation index uses the overall accuracy, which is the proportion of the number of correct pixels. The overall accuracy of the LU-3 model is 80.8%, and the overall accuracy of the LU-6 model is 81.52%. Without setting overlap region, the overall accuracy of the LU-6 model dropped to 81.23%. In this paper, the optimized classification method for remote sensing image that we proposed is better than the original DCNN in the field of computer vision from the perspective of qualitative and quantitative. Therefore, the result of the LU-6 model is finally adopted as a land-use classification product of the Guangdong Province of China. Considering that there are many errors in the classification label data, and the accuracy evaluation is based on this label data as the ground truth value, the accuracy of our deep learning model is actually higher than 81.52%.

## 4. CONCLUSIONS

In this study, we propose a fast and precise method for land-use classification of large-scale high spatial resolution satellite images at the provincial level. Compared with the traditional method, we complete a large-scale land-use classification in a small amount of time. We use a better data slicing method, which can eliminate the remote sensing edge effect, expand the data set, and suppress the abnormal features generated by the image boundary padding operation at the same time. We optimize the DCNN, and added more remote sensing data, which is beneficial to exploit more feature information. The method in this paper also has a certain tolerance for the error samples; there are still enough samples for training in case that parts of the sample label data is not accurate.